\pdfoutput=1

\documentclass[11pt]{article}

\usepackage{acl}

\usepackage{times}
\usepackage{latexsym}
\usepackage{booktabs}
\usepackage{amssymb}
\usepackage{amsmath}
\usepackage{multicol}
\usepackage{multirow}
\usepackage{comment}
\usepackage[T1]{fontenc}

\usepackage[utf8]{inputenc}

\usepackage{microtype}

\usepackage{inconsolata}

\usepackage{graphicx}

\usepackage[utf8]{inputenc}
\usepackage{tcolorbox}
\usepackage{enumitem}

\title{Improving Expert Radiology Report Summarization by Prompting Large Language Models with a Layperson Summary}

\author{Xingmeng Zhao, Tongnian Wang, \and Anthony Rios \\
  Department of Information Systems and Cyber Security \\ The University of Texas at San Antonio \\
  \texttt{\{xingmeng.zhao, tongnian.wang, anthony.rios\}@utsa.edu}}

\begin{document}
\maketitle
\begin{abstract}
Radiology report summarization (RRS) is crucial for patient care, requiring concise ``Impressions'' from detailed ``Findings.'' This paper introduces a novel prompting strategy to enhance RRS by first generating a layperson summary. This approach normalizes key observations and simplifies complex information using non-expert communication techniques inspired by doctor-patient interactions. Combined with few-shot in-context learning, this method improves the model's ability to link general terms to specific findings. We evaluate this approach on the MIMIC-CXR, CheXpert, and MIMIC-III datasets, benchmarking it against 7B/8B parameter state-of-the-art open-source large language models (LLMs) like Meta-Llama-3-8B-Instruct. Our results demonstrate improvements in summarization accuracy and accessibility, particularly in out-of-domain tests, with improvements as high as 5\% for some metrics.

\end{abstract}


\section{Introduction}
Radiology reports summarization (RRS) is an interesting task to explore natural language processing (NLP) methods in the biomedical domain from a computational perspective~\cite{van2023radadapt}. RRS involves generating concise ``Impressions'' from the detailed ``Findings'' and images in radiology reports. These reports, critical for patient diagnosis, treatment planning, and maintaining comprehensive records, are written by radiologists based on medical imaging techniques like X-rays, CT scans, MRI scans, and ultrasounds. The ``Findings'' section details objective observations from the imaging, while the ``Impressions'' section provides the radiologist's professional interpretation and diagnostic conclusions. 

In biomedical applications, the effectiveness of large language models (LLMs) models largely depends on their adaptation through domain- and task-specific fine-tuning~\cite{singhal2023large}. LLMs have shown remarkable proficiency in natural language understanding and generation, making them adaptable to various tasks. However, fine-tuning large models like GPT-3, with billions of parameters, requires substantial computational resources and high costs. To address these issues, researchers have shifted towards more efficient techniques like parameter-efficient fine-tuning (PEFT) and prompting~\cite{van2023radadapt, van2023clinical}, leveraging existing model capabilities while reducing computational demands~\cite{liu2022few}. 


In contrast, prompting through in-context learning (ICL)~\cite{brown2020language, dong2022survey} provides a practical alternative to extensive fine-tuning of LLMs. In ICL, relevant information is embedded directly within prompts, allowing LLMs to adapt to tasks with few-shot demonstrations~\cite{lampinen2022can} quickly. By carefully crafting these prompts, researchers can guide LLMs to generate accurate responses by providing clear context and examples. Techniques such as Retrieval-Augmented Generation~\cite{wang2023utsa} can further improve this process. Prompting has also proven effective in converting complex radiological data into clear and concise summaries~\cite{chen2022toward}. Moreover, \citet{nori2023can} found that combining ICL with explanations enhances the adaptation of general LLMs to specialized tasks, such as medical question answering, by integrating intermediate reasoning steps and thus improving problem-solving abilities~\cite{zhang2022automatic}. However, generating explanations for summarization tasks is inherently more challenging compared to question-answering and traditional text classification.

Moreover, LLMs trained on general text corpora often lack the specific knowledge required for specialized fields, limiting their performance~\cite{yao2023knowledge, holmes2023evaluating}. Addressing this deficiency typically involves extensive fine-tuning, which is resource-intensive and costly. While ICL can help by embedding relevant information within prompts, this alone is not always sufficient~\cite{brown2020language, dong2022survey}. Intuitively, non-fine-tuned models are ``non-experts'' in the medical domain, especially smaller open-source models.

However, in real-world settings (e.g., in actual doctor-patient conversations), research indicates that scientific or technical knowledge can be effectively transferred to non-experts through communication techniques like reformulation and simplification, which simplifies complex information and uses straightforward language to enhance understanding~\cite{gulich2003conversational}. Hence, inspired by effective doctor-patient communication methods, this paper proposes a novel prompting strategy that combines simplification techniques with ICL to enhance the performance of non-expert LLMs in specialized areas. This approach aims to improve model performance without needing costly fine-tuning~\cite{nori2023can, zhang2022automatic} by simplifying complex information and incorporating it through prompts before an expert summary is generated. The in-context examples have layperson/simplified language as part of them to help guide the model for a new example. From another perspective, we introduce a novel approach that first generates a layperson (non-expert) summary to \textit{normalize} key observations. Radiologists often have distinct reporting styles, leading to variations in terminology and impacting the consistency of medical documentation~\cite{yan2023style}. Additionally, the vast number of illnesses increases the variety of vocabulary encountered in reports. Normalizing terms in the layperson summary can better identify patterns between simplified summaries and detailed expert impressions, making it easier to link general terms to specific findings~\cite{peter2024simplicity}. For example, normalizing ``pneumonia'' and ``bronchitis'' to ``infection of the lungs'' helps the model recognize important concepts in the in-context examples, even if pneumonia is used in the test instance while bronchitis is used in the in-context examples. The LLM can then connect them back to the findings (summary).

\begin{figure*}[t]
    \centering
    \includegraphics[width=.75\linewidth]{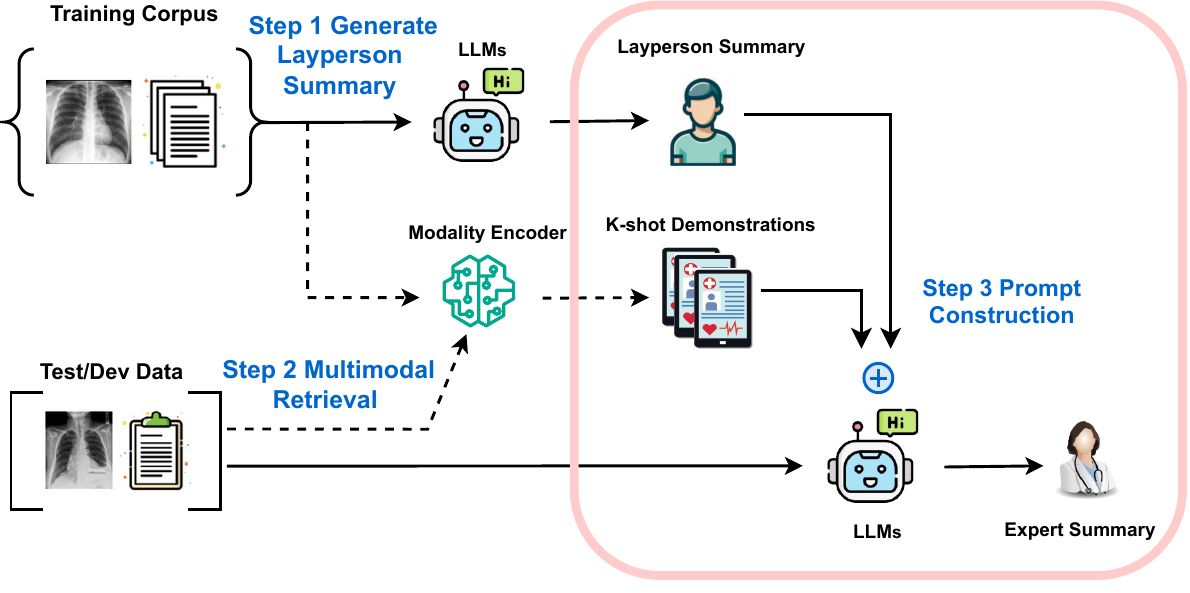}
    \caption{Overview of the LaypersonPrompt Framework. First, we generate layperson summaries from the training corpus using LLMs prompting. Then, for a test input, we use multimodal retrieval to find relevant examples. Finally, we incorporate these layperson summaries into the prompt, applying patient-doctor communication techniques to improve the model's reasoning.}
    \label{fig:layperson-pipeline} \vspace{-1em}
\end{figure*}

In summary, this paper makes the following contributions:
\vspace{-0.5em}
\begin{enumerate}
\item We introduce a novel prompting approach inspired by doctor-patient communication techniques that generate a simplified (layperson) summary before the expert summary. This strategy, combines with a few-shot ICL with the layperson summary, enhances RRS using non-expert LLMs.
\vspace{-0.5em}

\item We evaluate LLM performance on three RRS datasets: MIMIC-CXR~\cite{johnson2019mimic}, CheXpert~\cite{irvin2019chexpert}, and MIMIC-III~\cite{johnson2016mimic}, and benchmark against open-source LLMs like Meta-Llama-3-8B-Instruct~\cite{llama3modelcard} for comprehensive comparison.
\vspace{-0.5em}
\item We conduct a comprehensive analysis to determine the optimal modality for ICL. We also examine the required number of examples and the impact of layperson summaries on impressions and evaluate model performance on inputs of different lengths.\footnote{See the appendix for complete analysis.}

\end{enumerate}

\section{Related Work}
\noindent \textbf{LLMs for Medicine.} Recent advances in LLMs have demonstrated that LLMs can be adapted with minimal effort across various domains and tasks. These expressive and interactive models hold great promise due to their ability to learn broadly useful representations from the extensive knowledge encoded in medical corpora at scale~\cite{singhal2023large}. Fine-tuned general-purpose models have proven effective in clinical question-answering, protected health information de-identification~\cite{sarkar2024identification}, and relation extraction~\cite{hernandez2023we}. Some LLMs, such as BioGPT \cite{luo2022biogpt} and ClinicalT5 \cite{lu2022clinicalt5}, have been trained from scratch using clinical domain-specific notes, achieving promising performance on several tasks. Additionally, in-context learning with general LLMs like InstructGPT-3 ~\cite{ouyang2022training}, where no weights are modified, has shown good performance ~\cite{agrawal-etal-2022-large}. They have also demonstrated the ability to solve domain-specific tasks through zero-shot or few-shot prompting and have been applied to various medical tasks, such as medical report summarization~\cite{otmakhova-etal-2022-m3} and medical named entity recognition~\cite{hu2023zero}. But, this generally only works with closed-source models such as GPT4.

\vspace{2mm}
\noindent \textbf{Retrieval-Augmented LLMs.} Retrieval augmentation connects LLMs to external knowledge to mitigate factual inaccuracies. By incorporating a retrieval module, relevant passages are provided as context, enhancing the language model's predictions with factual information like common sense or real-time news \cite{ma-etal-2023-query}. Recent studies indicate that retrieval-augmented methods can enhance the reasoning ability of LLMs and make their responses more credible and traceable \cite{shi2023replug,yao2023react,nori2023can,ma-etal-2023-query}. For example, \citet{shi2023replug} trains a dense retrieval model to complement a frozen language model. By using feedback from the LLM as a training objective, the retrieval model is optimized to provide better contextual inputs for the LLM. \citet{yao2023react} focuses on designing interactions between the retriever and the reader, aiming to trigger emergent abilities through carefully crafted prompts or a sophisticated prompt pipeline. Our approach combines retrieval-augmented methods with layperson summaries to enhance general LLMs reasoning in radiology report summarization, using patient-doctor communication techniques for better understanding and accuracy.

\vspace{1.5mm}
\noindent \textbf{Communication Techniques for Laypersons.} Non-experts, such as patients, have been shown to perform well on expert tasks, like medical decision-making and understanding complex topics, when information is simplified using effective communication techniques~\cite{gulich2003conversational, leblanc2014patient, allen2023jargon, van2007patient, neiman2017cdc}. This simplification can also improve general LLMs performance on specialized tasks. Studies demonstrate that non-experts, with supervision, can generate high-quality data for machine learning, producing expert-quality annotations for tasks like identifying pathological patterns in CT lung scans and malware run-time similarity~\cite{o2017crowdsourcing, vanhoudnos2017malware, snow2008cheap}. Recent research has shown that LLMs can simplify complex medical documents, such as radiology reports, making them more accessible to laypersons. For instance, ChatGPT has been used to make radiology reports easier to understand, bridging the communication gap between medical professionals and patients~\cite{jeblick2023chatgpt, lyu2023translating, li2023decoding}. Inspired by these findings, we explore whether presenting expert-level information in simpler language can improve the performance of general LLMs on tasks that typically require specialized knowledge, such as those involving medical data.

\section{Methodology}

\textcolor{black}{In this section, we describe our prompting strategy. Figure~\ref{fig:layperson-pipeline} shows a high-level overview of our approach. Our strategy has three main components: 1) layperson summarization of the training dataset used as in-context examples; 2) ``multimodal demonstration retrieval,'' which is how we generate embeddings to find relevant in-context examples; and 3) final expert summary prompt construction, which is how we integrate the layperson summaries and in-context examples to generate the final expert summary. We describe each component in the following subsections and how the three components are integrated into a unified prompt.}

\vspace{2mm}
\noindent \textbf{Step 1: Layperson Summarization of the Training Dataset.} \textcolor{black}{Layperson summarization involves converting complex medical texts into more straightforward language, enhancing accessibility and understanding for individuals without medical expertise~\cite{cao2020expertise}. For instance, rephrasing ``pulmonary edema'' as ``fluid in the lungs'' makes it more comprehensible. This approach not only helps to bridge the knowledge gap for laypeople but also plays an important role in helping models better understand and summarize medical content. Intuitively, by generating simplified summaries as an intermediate step, models can more effectively capture the semantic meaning of the texts~\cite{liu2024improving, sulem2018simple,paetzold2016unsupervised,shardlow2019neural}. In this context, we generate layperson summaries as an intermediate step for all training examples to enhance the generation of expert summaries.}




\begin{figure}[t]
    \centering
    \includegraphics[width=\linewidth]{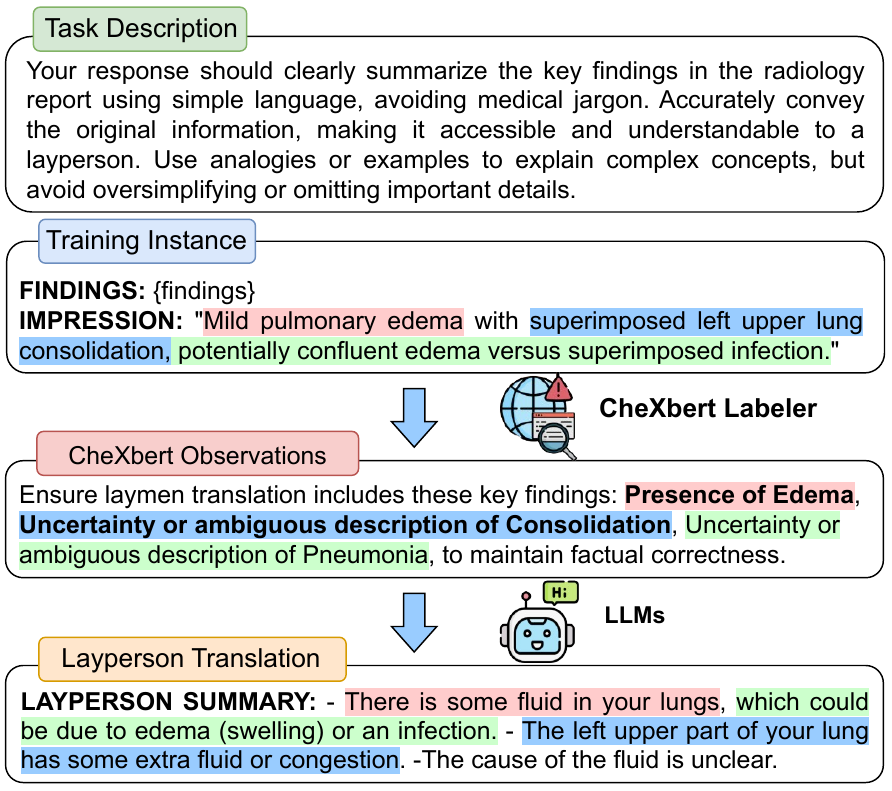}
    \caption{ \textcolor{black}{Step 1: Layperson Summarization of the Training Dataset. An illustration of the layperson summary prompt used to generate layperson summaries for training examples.} Disease observations are highlighted in different colors. The illustration shows a single example, with Instruction and Response sections repeated multiple times using few-shot in-context examples.}\vspace{-1em}
    \label{fig:translation}
\end{figure}

\textcolor{black}{To generate accurate layperson summaries, we use a zero-shot prompting strategy enhanced with metadata from an external tool. Specifically, we employ the CheXbert labeler~\cite{smit2020combining} to extract important medical observations from radiology impressions (e.g., ``No Finding'', ``Pneumonia'', ``Cardiomegaly'', etc.). Using CheXbert's output, we create additional instructions for the language model that include these key concepts. The exact form of the prompt is shown in Figure~\ref{fig:translation}. This prompt integrates the Task Instruction, Findings, Impression, and the extracted CheXbert concepts. We then use this prompt to generate layperson summaries and store these summaries along with their corresponding Findings and Impressions as training triples, which are used as in-context examples.}

\vspace{2mm}
\noindent \textbf{Step 2: Multimodal Demonstration Retrieval} A major feature of our system is finding similar examples in the training dataset for each test example to use as in-context examples. In our approach, we focus on substantially improving the performance of LLMs with a few well-chosen examples to generate more accurate and standardized summaries. Selecting the right examples is a critical task in few-shot learning, as it greatly affects the effectiveness of the LLMs. To ensure the selection of the most relevant examples, we employ multimodal retrieval strategy based on our prior work for fine-tuning-based methods~\cite{wang2023utsa}. In this framework, we retrieve the top-$k$ similar radiology report based on different modalities, i.e., chest X-ray images, text findings, and multi-modal data (combining findings and images) from a medical corpus using a pre-trained multi-modal encoder. Then, we include the findings and impressions of the top $k$ of the most similar report as input in our final prompt.

\begin{figure}[t]
    \centering
    \includegraphics[width=\linewidth]{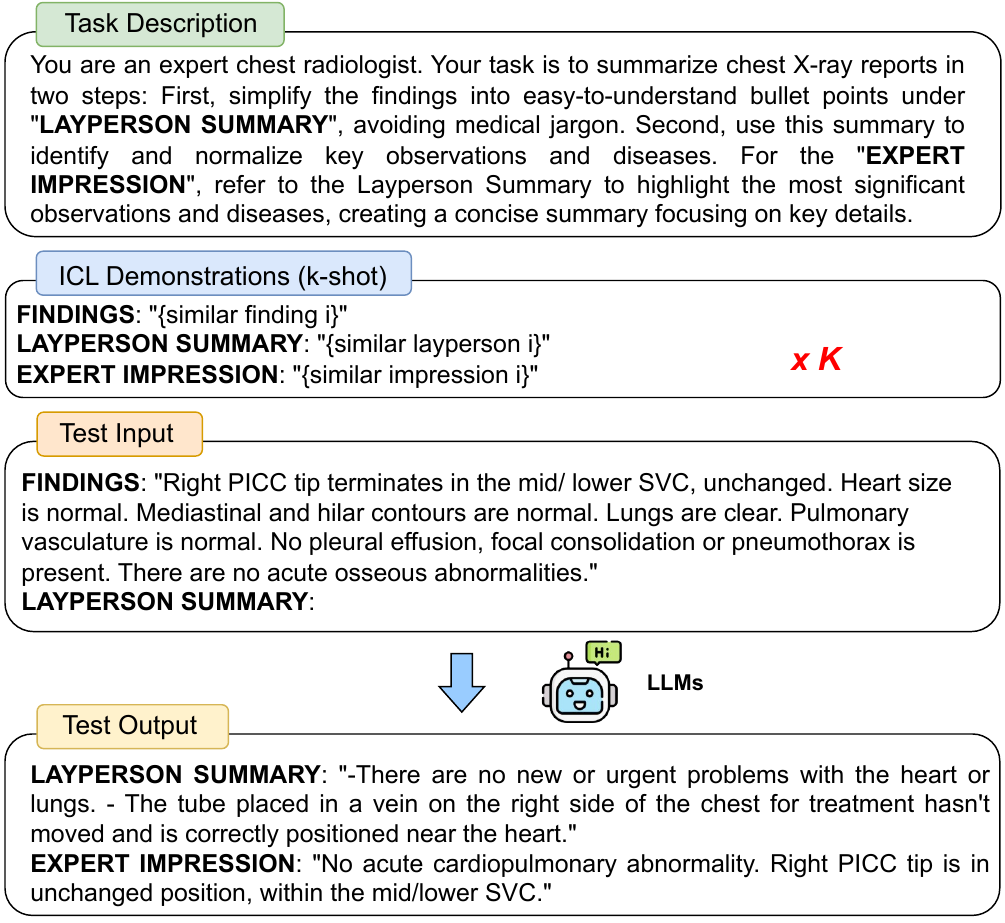}
    \caption{\textcolor{black}{Step 3: Final Expert Summary Prompt Construction. Example of LaypersonPrompt. This is the final prompt after finding in-context examples to generate the final expert summary (i.e., the Impression section).}}
    \label{fig:layperson-prompt}
\end{figure}

Formally, given an input instance $x_i$ consisting of a text input $w$ and image $m$, our goal is to retrieve the most similar examples $\{x_1, \ldots, x_{\mathcal{N}(x_i)}\}$, where $\mathcal{N}(x_i)$ represents the top $k$ indexes for similar examples to $x_i$. To achieve this, we employ a multimodal image-text retrieval model that uses separate encoders for text and image modalities alongside a multimodal encoder for integrating their embeddings. Specifically, the image is processed through a pre-trained Vision Transformer (ViT) model~\citep{dosovitskiy2020image} to generate image embeddings. Since some findings correspond to multiple images, we average all image embeddings corresponding to the same findings. Next, we adapt a pre-trained Transformer encoder-decoder model, such as Clinical-T5 \citep{lehmanclinical}, to handle multimodal inputs. Specifically, we pass the findings as input to the T5 encoder and initialize its hidden state with the averaged image embeddings. The final $EOS$ token from the T5 encoder is used as the multimodal embeddings. Note that this model cannot be used as-is with the initial pre-trained models. Instead, we train this model where the T5 encoder outputs are passed to the T5 decoder to generate the impressions. After training the joint model, we remove the decoder and only the embeddings will be used later.

\begin{table*}[t]
\resizebox{\linewidth}{!}{
\begin{tabular}{@{}llrrrrr@{}}
\toprule
 &  & \textbf{BLEU4} & \textbf{ROUGEL} & \textbf{BERTScore} & \textbf{F1-cheXbert} & \textbf{F1-RadGraph} \\ 
 \midrule
\multirow{3}{*}{Zero-Shot}      
& OpenChat-3.5-7B  & 3.98 & 21.74 & 42.74 & 64.98 & 18.34 \\
& Starling-LM-7B  & 3.64 & 21.28 & 42.29 & 64.63 & 17.93 \\ 
& Meta-Llama-3-8B-Instruct &  5.19 & 23.56 & 40.99 & 66.65 &23.56\\
\midrule
\multirow{3}{*}{Few-Shot} 
& OpenChat-3.5-7B  & 8.24 & 27.44 & 45.86 & 67.00 & 26.90 \\
& Starling-LM-7B  & 6.79 & 25.85 & 44.71 & 66.76 & 25.13 \\ 
& Meta-Llama-3-8B-Instruct & 6.33 & 25.81 & 40.19 & 65.72 & 30.13\\
\midrule
\multirow{3}{*}{Few-Shot + Chexbert} 
& OpenChat-3.5-7B  &  8.11 & 27.68 & 44.62 & 65.71 & 26.80 \\
& Starling-LM-7B  & 6.29 & 25.57 & 42.96 & 63.56 & 24.09\\ 
& Meta-Llama-3-8B-Instruct &  9.20 &  28.25 &  44.63 & 67.23 & \textbf{30.48}\\
\midrule
\multirow{3}{*}{Few-Shot + Layperson} 
& OpenChat-3.5-7B  & 8.96 & 28.46 & 45.35 & 67.00 & 27.90 \\
& Starling-LM-7B  & 8.35 & 26.97 & 44.93 & 66.29 & 26.94 \\ 
& Meta-Llama-3-8B-Instruct &  \textbf{9.36} & \textbf{29.03} & \textbf{46.91} & \textbf{68.64} & 29.96 \\
\bottomrule
\end{tabular}}
\caption{Overall performance across the four prompts on the MIMIC CXR in-domain test.}
\label{tab:test-results}
\end{table*}

\begin{table*}[t]
\resizebox{\linewidth}{!}{
\begin{tabular}{@{}llrrrrr@{}}
\toprule
 &  & \textbf{BLEU4} & \textbf{ROUGEL} & \textbf{BERTScore} & \textbf{F1-cheXbert} & \textbf{F1-RadGraph} \\
 \midrule
\multirow{3}{*}{Zero-Shot} 
& OpenChat-3.5-7B & 2.22 & 25.14 & 47.10 & 68.95 & 10.68 \\
& Starling-LM-7B & 2.18 & 24.61 & 46.49 & 70.36 & 10.50 \\
& Meta-Llama-3-8B-Instruct &  2.01 & 23.69 & 42.53 & 68.76 & 10.99\\
\midrule
\multirow{3}{*}{Few-Shot} 
& OpenChat-3.5-7B & 5.23 & 27.43 & 48.00 & 70.32 & 12.35 \\
& Starling-LM-7B & 4.66 & 26.77 & 47.20 & 70.68 & 11.64 \\
& Meta-Llama-3-8B-Instruct &  3.37 & 22.09 & 39.35 & 66.49 & 11.22 \\
\midrule
\multirow{3}{*}{Few-Shot + Chexbert} 
& OpenChat-3.5-7B &  5.43 & 26.50 & 44.95 & 69.80 & 12.31 \\
& Starling-LM-7B &  3.40 & 23.93 & 44.12 & 64.90 & 10.74 \\
& Meta-Llama-3-8B-Instruct &  3.79 & 24.75 & 42.52 & 70.05 & 11.79 \\
\midrule
\multirow{3}{*}{Few-Shot + Layperson} 
& OpenChat-3.5-7B & \textbf{7.74} & 28.71 & \textbf{48.04} & 71.28 & 13.15 \\
& Starling-LM-7B & 7.01 & 28.90 & 48.02 & 71.02 & 12.93 \\
& Meta-Llama-3-8B-Instruct &  7.47 & \textbf{29.03} & 47.29 & \textbf{71.91} & \textbf{13.63}\\
\bottomrule
\end{tabular}}
\caption{Overall performance across the four prompts on the Stanford Hospital (out-of-domain) test set. The in-context examples for this dataset are from the MIMIC-CXR dataset.} \vspace{-1em}
\label{tab:hidden-results}
\end{table*}

\begin{table*}[t]
\resizebox{\linewidth}{!}{
\begin{tabular}{@{}llrrrrr@{}}
\toprule
 &  & \textbf{BLEU4} & \textbf{ROUGEL} & \textbf{BERTScore} & \textbf{F1-cheXbert} & \textbf{F1-RadGraph} \\
 \midrule
\multirow{3}{*}{Zero-Shot} 
& OpenChat-3.5-7B  & 4.61  & 19.85  & 43.02  & 52.06  & 21.41  \\
& Starling-LM-7B  & 4.51  & 19.52  & 42.57  & 51.77  & 21.19 \\ 
& Meta-Llama-3-8B-Instruct &  5.61 & 20.34 & 41.70 & 51.87 & 24.43 \\
\midrule
\multirow{3}{*}{Few-Shot}          
& OpenChat-3.5-7B  & 8.02  & 22.33  & 45.56  & 52.71  & 23.37 \\
& Starling-LM-7B  & 7.95  & 21.98  & 45.05  & 52.49  & 23.01\\ 
& Meta-Llama-3-8B-Instruct & 6.25 & 20.03 & 38.75 & 47.54 & 24.76\\
\midrule
\multirow{3}{*}{Few-Shot + Chexbert} 
& OpenChat-3.5-7B  & 8.05 & 21.94 & \textbf{45.61} & 51.03 & 24.70\\
& Starling-LM-7B   & 9.28 & 22.43 & 44.93 & 49.94 & 22.05\\ 
& Meta-Llama-3-8B-Instruct &  7.39 & 21.36 & 40.76 & 48.06 & 24.40 \\
\midrule
\multirow{3}{*}{Few-Shot + Layperson} 
& OpenChat-3.5-7B   & 8.62  & \textbf{22.95}  & 45.31  & \textbf{52.81}  &  24.37 \\
& Starling-LM-7B   &10.02 &  22.70 & 45.14  &  51.83 & 24.32 \\ 
& Meta-Llama-3-8B-Instruct & \textbf{10.03} & 21.49 & 45.29 & 50.78 & \textbf{24.99}\\
\bottomrule
\end{tabular}}
\caption{Overall performance across the four prompts on the MIMIC III.} \vspace{-1em}
\label{tab:mimiciii-results}
\end{table*}

\vspace{2mm}
\noindent \textbf{Step 3: Expert Summary Prompt Construction} The final step in our pipeline involves prompting an LLM to generate an expert summary, following the generation of layperson summaries for all training examples and identifying relevant in-context examples for development/test instances using multimodal demonstration retrieval. The prompt comprises three main components: 1) Task Instruction; 2) In-context learning examples (ICL Demonstrations); and 3) the test input instance. An example is shown in Figure~\ref{fig:layperson-prompt}.

First, the Task Instruction specifies that the model should create a layperson summary followed by an expert impression. Detailed guidelines are provided for generating both the layperson summary and the expert impression. It is important to note that the layperson summary is generated as part of this prompt for the input instance before generating the expert impression. The prompt defined in Step 1 is only used for the training examples. Next, given the input instance's Findings text and radiology image, we use the same multi-modal encoder and retrieval approach described in Step 2 to find relevant in-context examples from the training dataset.  We generate a sequence of up to 32 in-context demonstrations. After identifying the relevant training examples, we append each training instance's Findings, layperson summary, and Impression to generate the sequence of in-context examples. Finally, we append the Findings section of the text instance and the string ``Layperson Summary:''. The model will first generate the layperson summary followed by the final expert Impression.

Why does generating a layperson summary before the expert impression work? Models can produce general information (e.g., ``Infection of the lungs'' for ``pneumonia'') in the layperson summary, which helps to standardize the content in the Findings before creating the Impression. This means different illnesses can be simplified to the same concept (e.g., ``bronchitis'' can also be simplified to ``Infection of the lungs''). The idea is that the model can find common patterns in these general (layperson) expressions that correlate with the expert Impression, as long as the Findings have similar content. After generating the layperson summary, the model only needs to connect the general terms in the summary to the specific details in the Findings, similar to coreference resolution. Without the layperson summary, the model must directly find patterns in the more varied Findings section, making the task more complex.


\section{Experimental Results}

This section covers the datasets, evaluation metrics, overall results, and error analysis. 

\vspace{1mm}
\noindent \textbf{Datasets} In this study, we evaluate our prompting method on two radiology reports summarization datasets. The MIMIC-III summarization dataset, as introduced by~\cite{johnson2016mimic, chen-etal-2023-toward}, contains 11 anatomy-modality pairs (i.e., 11 body parts and imaging modalities such as head-MRI and abdomen-CT). The dataset consists of train, validation, and test splits of 59,320, 7,413, and 6,531 findings-impression pairs, respectively. The MIMIC-III dataset only contains radiology reports without the original images. On the other hand, the MIMIC-CXR summarization dataset~\cite{johnson2019mimic} is a multimodal summarization dataset containing findings and impressions from chest X-ray studies and corresponding chest X-ray images. It comprises 125,417 training samples, 991 validation samples, and 1624 test samples. Additionally, we incorporate an out-of-institution multimodal test set of 1000 samples from the Stanford hospital~\cite{irvin2019chexpert} to assess the out-of-domain generalization of models trained on MIMIC-CXR. We use OpenChat-3.5-7B~\cite{wang2023openchat}, Starling-LM-7B~\cite{starling2023}, and Meta-Llama-3-8B-Instruct~\cite{llama3modelcard} in our experiments to compare model performance.


\vspace{1mm}
\noindent \textbf{Evaluation Metrics.} Performance is evaluated using the following metrics: BLEU4~\cite{papineni2002bleu}, ROUGE-L~\cite{lin2004rouge}, Bertscore~\cite{Zhang2020BERTScore}, F1CheXbert~\cite{delbrouck-etal-2022-vilmedic}, and F1RadGraph~\cite{delbrouck-etal-2022-improving}. Intuitively, BLEU4 measures the precision, while ROUGE-L assesses the recall of the n-gram overlap between the generated radiology reports and the original summaries. BERTScore calculates the semantic similarity between tokens of the reference summary and the hypothesis, where the hypothesis refers to the model-generated summary. F1CheXbert uses CheXbert~\cite{smit-etal-2020-combining}, a Transformer-based model, to evaluate the clinical accuracy of generated summaries by comparing identified chest X-ray abnormalities in the generated reports to those in the reference reports. F1RadGraph, an F1-score style metric, leverages the RadGraph~\cite{jain2021radgraph} annotation scheme to evaluate the consistency and completeness of the generated reports by comparing them to reference reports based on observation and anatomy entities.

\vspace{2mm}
\noindent \textbf{Overall Results.} Table~\ref{tab:test-results} show the performance of Zero-Shot prompting, Few-Shot prompting, Few-Shot + Chexbert prompting, and our Few-Shot + Layperson prompting strategies for the radiology reports summarization task on the MIMIC-CXR dataset. The Few-Shot + Chexbert method adds disease keywords to help the model focus. In contrast, the Few-Shot + Layperson method mimics doctor-patient communication by creating a simplified summary for laypeople before generating the expert summary. We find that the Few-Shot + Layperson method yielded the best results overall. Meta-Llama-3-8B-Instruct achieved the highest scores in BLEU4 (9.36), ROUGEL (29.03), BERTScore (46.91), and F1-cheXbert (68.64), and strong performance in F1-RadGraph (29.96). OpenChat-3.5-7B and Starling-LM-7B also showed significant improvements with Few-Shot + Layperson, notably in BLEU4 and F1-RadGraph. Specifically, on OpenChat-3.5-7B, ROUGE-L, and F1-RadGraph, there were respective increases of 0.78 and 1.10 compared to not using the layperson summary. For Starling-LM-7B, these metrics rise by 1.12 and 2.85, respectively. These results suggest incorporating a layperson summary can enhance model performance in summarizing radiology reports.

On the Stanford Hospital test set in Table~\ref{tab:hidden-results}, the Few-Shot + Layperson prompting yields a respective increase in performance across multiple metrics. OpenChat-3.5-7B achieved the highest BLEU4 (7.74) and BERTScore (48.04), while Meta-Llama-3-8B-Instruct led in ROUGEL (29.03), F1-cheXbert (71.91), and F1-RadGraph (13.63). Starling-LM-7B also showed substantial improvements in ROUGEL (28.90 vs. 23.93) and BERTScore (48.02 vs. 44.12) compared to Few-Shot + Chexbert. These results highlight the effectiveness of using a layperson summary to enhance model performance in summarizing radiology reports on the out-of-domain dataset.

The results of the comparison on the MIMIC-III dataset are detailed in Table~\ref{tab:mimiciii-results}. Our model demonstrates robust performance, indicating its capability to generalize across varied medical datasets. Specifically, Meta-Llama-3-8B-Instruct saw increases in BLEU4 (10.03 vs. 7.39) and F1-RadGraph (24.99 vs. 24.40) compared to Few-Shot + Chexbert. In summary, across all three datasets, it is evident that the Few-Shot + Layperson method shows noticeable improvements, especially on the out-of-domain test set. Incorporating an intermediate layperson summary, which mimics patient-doctor communication, introduces a step for ``easy-to-hard'' reasoning. This approach enhances the model's accuracy and its ability to generalize across different datasets in medical imaging and report summarization.

\begin{figure}[th]
    \centering
    \includegraphics[width=.75\linewidth]{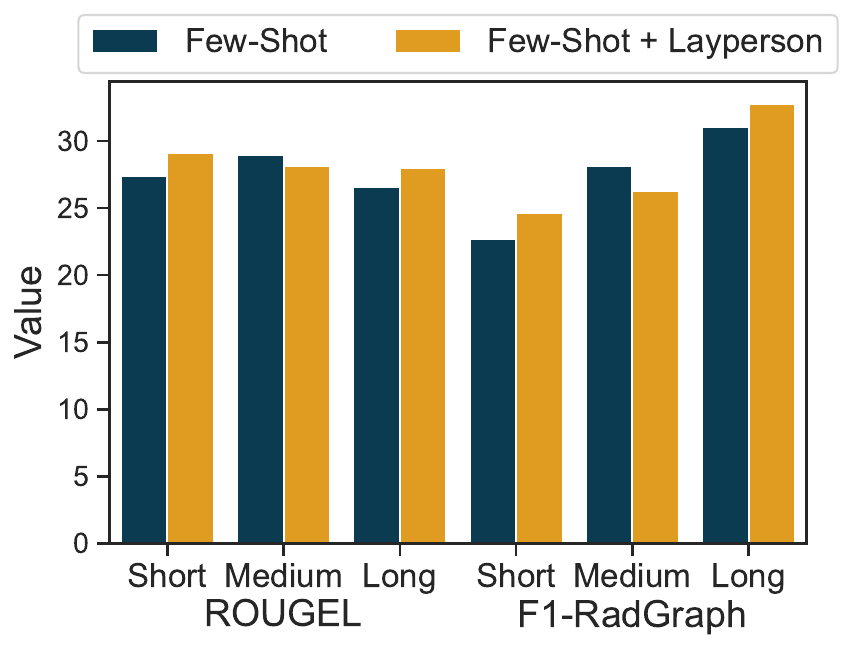}
\vspace{-0.5em}
\caption{Error Analysis on MIMIC-CXR Test Dataset: Performance Comparison of OpenChat-3.5-7B Model across Different Impression Lengths.
\label{fig:error}}
\vspace{-2em}
\end{figure}

\vspace{2mm}
\noindent \textbf{Error Analysis.} We conducted an error analysis of the OpenChat-3.5-7B model on the MIMIC-CXR test dataset, comparing the Few-Shot + Layperson prompting strategy to Few-Shot prompting using multimodal embeddings. We analyzed performance trends across different impression lengths using ROUGE-L for text similarity and F1-RadGraph for entity accuracy and completeness of the generated radiology reports. The results are shown in Figure~\ref{fig:error}. We found that ROUGE-L scores decrease with longer impressions while F1-RadGraph scores increase. This suggests that while the model's text similarity drops with longer impressions, its accuracy in identifying specific medical entities improves. The F1-RadGraph metric benefits from the richer context and greater detail in longer impressions, allowing the model to recognize better and generate relevant diseases or observations. Our layperson summary approach normalizes terminology and links general terms to specific findings, enhancing the model's accuracy. This approach is similar to how doctors explain complex medical information to patients: by simplifying and clarifying terms, they ensure understanding and accuracy. This technique helps the model identify and generate important entities better, leading to higher F1-RadGraph scores for longer impressions.

Conversely, the decrease in ROUGE-L scores for longer impressions may be due to the increased complexity and diversity of information. Longer impressions often include more nuanced and varied details, which may not exactly match the ground truth in phrasing or specific word choice. As a result, the text similarity measured by ROUGE-L tends to be lower for longer impressions. This metric is sensitive to the exact wording and word sequence, so the greater length and complexity can result in more divergence from the reference text.

Interestingly, when evaluated with the ROUGE-L metric, the Few-Shot + Layperson prompting approach performs best on shorter texts (29.11 vs. 27.41). For example, consider the short impression:
\begin{center}
\vspace{-.25em}
\tcbset{
    colframe=black,
    colback=white,
    boxrule=0.5mm,
    arc=3mm,
    width=1\linewidth,
    boxsep=5pt,
    left=5pt,
    right=5pt,
    top=5pt,
    bottom=5pt
}
\begin{tcolorbox}
    \textbf{Ground Truth Short Impression}: ``No acute cardiopulmonary process.''\\[.25em]
    \textbf{Layperson Summary}: ``The lungs look normal with no issues like infection or fluid build-up. The heart and nearby structures are normal, and there are no broken bones.''\\[.25em]
    \textbf{Few-Shot}: ``No acute thoracic pathology.''\\[.25em]
    \textbf{Few-Shot + Layperson (Our)}: ``No acute cardiopulmonary process.''
\end{tcolorbox}
\vspace{-.25em}
\end{center}
The example above provides a qualitative analysis of the models’ performance, with given findings. The Few-Shot method summarized the findings as ``No acute thoracic pathology,'' which is too broad compared to the precise ground truth ``No acute cardiopulmonary process.'' However, our method accurately matched the ground truth by first creating a layperson summary and then summarizing it, ensuring clarity and precision. This approach's effectiveness in capturing short, concise statements makes it better at closely matching the ground truth.


Despite the increasing complexity, the Few-Shot + Layperson prompting continues to outperform Few-Shot prompting in longer impressions. An example of a long impression is:
\begin{center}
\vspace{-.25em}
\tcbset{
    colframe=black,
    colback=white,
    boxrule=0.5mm,
    arc=3mm,
    width=1\linewidth,
    boxsep=5pt,
    left=5pt,
    right=5pt,
    top=5pt,
    bottom=5pt
}
\begin{tcolorbox}
    \textbf{Ground Truth Long Impression}: ``Right basilar opacity is probably atelectasis, but could represent early or developing pneumonia in the appropriate clinical setting.''\\[.25em]
    \textbf{Layperson Summary}: ``The right lower part of the lung has a subtle cloudy area, which might be a small lung collapse but could also be early or developing pneumonia, depending on the patient's condition.''\\[.25em]
    \textbf{Few-Shot}: ``Normal cardiomegaly with subtle lung base opacity, possibly atelectasis or early pneumonia.''\\[.25em]
    \textbf{Few-Shot + Layperson (Our)}: ``1. Subtle opacity at the right lung base, possibly representing atelectasis, but could be early or developing pneumonia in the appropriate clinical setting.''
\end{tcolorbox}
\vspace{-.25em}
\end{center}

For long impression, the Few-Shot method is less precise than the ground truth, adding unnecessary details like ``Normal cardiomegaly'' and missing the position terms ``right''. In contrast, our method simplifies complex findings into layperson terms and then translates them back into accurate expert summaries. For example, "Right basilar opacity is probably atelectasis, ... early or developing pneumonia" becomes "The right lower lung looks cloudy, likely a small collapse or early pneumonia." This layperson summary is then accurately converted to "Subtle opacity at the right lung base, possibly atelectasis or early pneumonia," ensuring clarity and precision. The improvement with longer texts is likely due to the extra context they provide, similar to detailed doctor-patient explanations. 


\section{Conclusion}



This paper introduces a novel prompting approach inspired by doctor-patient communication techniques. By first generating a simplified (layperson) summary before creating the expert summary and combining this with few-shot in-context learning, we aim to improve the summarization of radiology reports using general LLMs. Evaluations across three datasets (MIMIC-CXR, CheXpert, and MIMIC-III) show that this method improves performance, especially in out-of-domain tests.

However, this approach faces challenges due to the computational demands and context token limitations of the 7B model, particularly with longer, more complex medical reports. Future work will focus on optimizing token usage within these constraints and exploring larger models with expanded context capacities. By leveraging the principles of effective doctor-patient communication, our method aims to enhance non-expert LLMs performance in specialized fields without requiring extensive fine-tuning.

\section{Limitation}
While our approach shows improvements in radiology report summarization (RRS), several limitations must be considered. First, the evaluation metrics used, such as ROUGE-L, do not always correlate well with human evaluations, necessitating cautious interpretation of the results~\cite{wang2024llm}. Our study primarily relies on these automated metrics, which can overlook important nuances that human experts might catch. The absence of comprehensive human evaluations further limits the assessment of practical effectiveness. Incorporating detailed evaluations by human experts is crucial for accurately measuring model performance in real-world clinical settings in future research, as human assessments provide insights into the clinical relevance and accuracy of summaries that automated metrics may miss. 

Additionally, the use of 7B parameter open-source models may not be optimal. More powerful closed models, like GPT-4~\cite{achiam2023gpt} and Gemini~\cite{team2023gemini}, often perform better in summarization tasks. Including results from these advanced models could provide a more comprehensive comparison and potentially challenge the necessity of the intermediate layperson summary step. Furthermore, the computational demands and context token limitations of the 7B model present significant challenges, particularly with longer and more complex medical reports. This restricts the model's ability to process extensive and detailed information effectively. Differences in the quality and consistency of radiology reports from different datasets can also affect performance due to inconsistencies in terminology and reporting styles. Moreover, the current interaction between humans and non-expert LLMs can be improved. Incorporating communication techniques similar to doctor-patient interactions will enhance the human-AI experience by making complex information more accessible and understandable. This improvement aims to make LLMs more practical and effective for expert-level tasks in various areas, bridging the gap between specialized knowledge and everyday understanding.

\section{Ethics Statement}
In this work, we have introduced our Layperson Summary Prompting strategy, inspired by doctor-patient communication techniques. This approach aims to simplify complex medical findings into layperson summary first, then uses this simplified information to generate accurate expert summaries. However, it is important to address the ethical implications of using LLMs in this context. LLMs used for radiology report summarization can produce errors or biased outputs if the training data is of low quality or representative. These models also can be wrong, and such biases can lead to unfair outcomes and exacerbate health disparities. Therefore, radiologists should use AI-generated summaries as supportive tools, retaining control over clinical decisions. AI should be seen as an information resource to reduce time and cognitive effort, aiding in information retrieval and summarization, rather than as an interpretative agent providing clinical decisions or treatment recommendations. 

Additionally, integrating AI into clinical practice raises significant ethical considerations regarding patient privacy, data security, and informed consent. Using large volumes of sensitive patient data for training AI models necessitates stringent measures to protect patient rights and ensure data confidentiality. Ethical principles such as fairness, accountability, and transparency should guide the deployment of AI technologies in healthcare. These principles help ensure that AI systems are used responsibly and that the benefits of AI are distributed equitably among all stakeholders. Furthermore, potential risks associated with AI implementation include perpetuating existing biases, privacy breaches, and the misuse of AI-generated data, necessitating careful consideration and proactive management~\cite{yildirim2024multimodal}.

\bibliography{custom}

\appendix

\section{Appendix}
\label{sec:appendix}

\subsection{Baseline and Implementation Details}
For our baseline approach, we adopt a prefixed zero-shot prompting strategy~\cite{duan2019zero,zhao2021discrete}, which prepended a brief instruction to the beginning of a standard null prompt. We use the instruction, ``You are an expert chest radiologist. Your task is to summarize the radiology report findings into an impression with minimal text''. This instruction provides the model with a fundamental context for the RRS task. Immediately following the instruction, we append the specific findings from the report and then prompt the model with ``IMPRESSION:'' to initiate the generation process. Additionally, we investigate the effectiveness of few-shot ICL prompts with up to 32 similar examples, using the same template as our Few-Shot prompting method, which is not incorporating the intermediate reasoning step (i.e., without the layperson summary). 

\textcolor{black}{We conduct experiments with six open-source LLMs: OpenChat-3.5-7B~\cite{wang2023openchat}, Starling-LM-7B~\cite{starling2023},  Meta-Llama-3-8B-Instruct~\cite{llama3modelcard}, LLaMA-2-7B~\cite{touvron2023llama},  LLaMA-2-13B~\cite{touvron2023llama}, and GPT-Neo-2.7B~\cite{gao2020pile}. All experiments were conducted using two Nvidia A6000 GPUs. For the few-shot model, the average running time is around 2 hours. In contrast, the Few-Shot + Layperson models have an average running time of around 8 hours. Processing the MIMIC data with 24 examples takes approximately 36 hours. In our work, all of these models have been implemented using the Hugging Face framework~\cite{wolf2019huggingface}. Specifically, for the OpenChat-3.5-7B, Starling-LM-7B, and Meta-Llama-3-8B-Instruct are reported to have strong performance in common sense reasoning and problem-solving ability~\cite{starling2023}. OpenChat-3.5-7B is built on the Mistral 7B with conditioned reinforcement learning fine-tuning, and Starling-LM-7B is built on OpenChat-3.5-7B with reinforcement learning from AI feedback. Moreover, OpenChat-3.5-7B, Starling-LM-7B and Meta-Llama-3-8B-Instruct had sufficient token maximums (8,192) compared to other 7B models (e.g., LLaMA-2-7B and  LLaMA-2-13B only has 4096 maximum tokens, GPT-Neo-2.7B has 2048 maximum tokens).} To select the best parameters in our study, we employed ROUGE-L and F1RadGraph metrics on the validation set. These metrics help determine the most effective parameter settings for the model. The ROUGE-L metric focuses on the longest common subsequence and is particularly suitable for evaluating the quality of text summaries. On the other hand, the F1RadGraph is specifically designed to assess the accuracy of extracting and summarizing key information from radiology reports by analyzing entity similarities. 

For optimizing our model's hyper-parameters, we employed a random search strategy. This involved experimenting with various settings: the number of prepended similar examples was varied across a set {2, 8, 12, 16, 24, 32}, and these examples were matched using different modality embeddings (text, image, or multimodal), all while employing the same template. We find that for the OpenChat-3.5-7B model and Meta-Llama-3-8B-Instruct, the best performance is achieved with 32 examples for \textcolor{black}{both Few-Shot and Few-Shot + Layperson prompting methods}. In contrast, the Starling-LM-7B model exhibits optimal performance with 32 examples when using the \textcolor{black}{Few-Shot} prompt and 24 examples for the \textcolor{black}{Few-Shot + Layperson prompt}. Additionally, we experimented with temperature settings ranging from 0.1 to 0.9, top p values set between 0.1 and 0.6, and top k values of 10, 20, and 30. Through this exploratory process, we identified the most effective settings as a temperature of 0.2, a top p value of 0.5, and a top k setting of 20. We adopt the same hyperparameters for all experiments. These settings yielded the best results in our evaluations. It's significant to note the impact of the ``temperature'' parameter on the diversity of the model's outputs. Higher temperature values add more variation, introducing a greater level of randomness into the content generated. This aspect is especially valuable for adjusting the output to meet specific requirements for creativity or diversity.

To ensure compatibility with the model's capabilities, we restricted the length of the prompt (which includes the instruction, input, and output instance) to 7800 tokens. This limit was set to prevent exceeding the model's maximum sequence length of 8,192 tokens for OpenChat-3.5-7B, Starling-LM-7B model and Meta-Llama-3-8B-Instruct. \textcolor{black}{For LLaMA-2-7B and LLaMA-2-13B models, we constrain the prompt length to 3800 tokens, and for GPT-Neo-2.7B, it is set to 1700 tokens.} In cases where prompts exceeded this length, they were truncated from the beginning, ensuring that essential information and current findings were preserved. Moreover, we constrained the generated output to a maximum of 256 tokens to strike a balance between providing detailed content and adhering to the model's constraints. This approach was key in optimizing the effectiveness of summarization within the operational limits of the 7B models.

\subsection{Discussion and Model Analysis} 

\begin{figure*}[htbp]
\centering
\includegraphics[width=\textwidth]{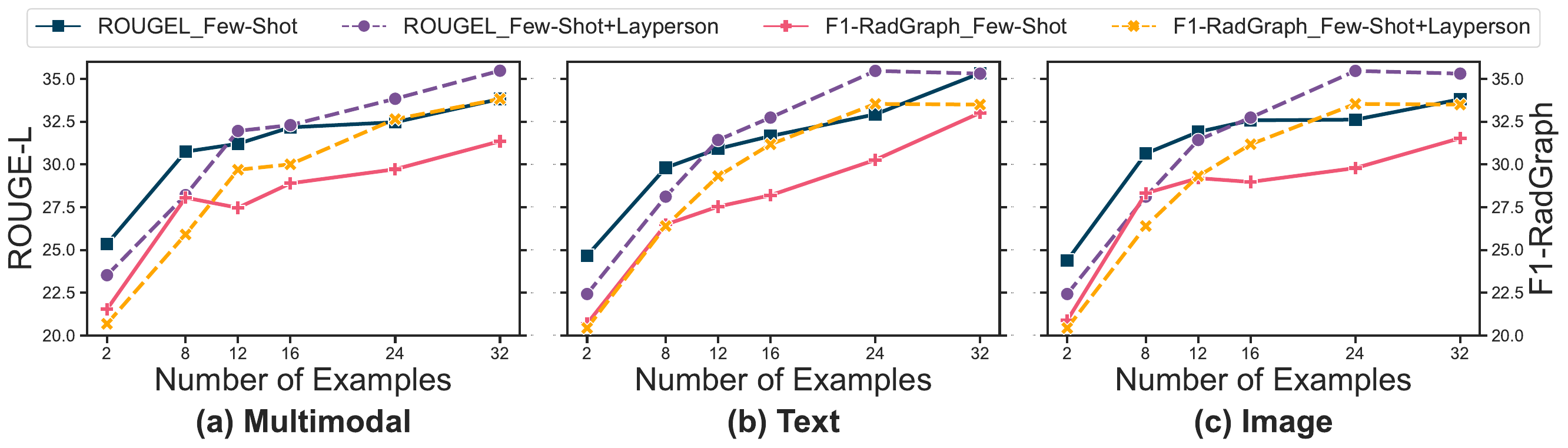}\par
\vspace{-0.5em}
\caption{Validation results vs. the number of in-context examples across various prompt types and modality embeddings on OpenChat-3.5-7B.
\label{fig:openchat-plot}}
\end{figure*}

\begin{figure*}[htbp]
\centering
\includegraphics[width=\textwidth]{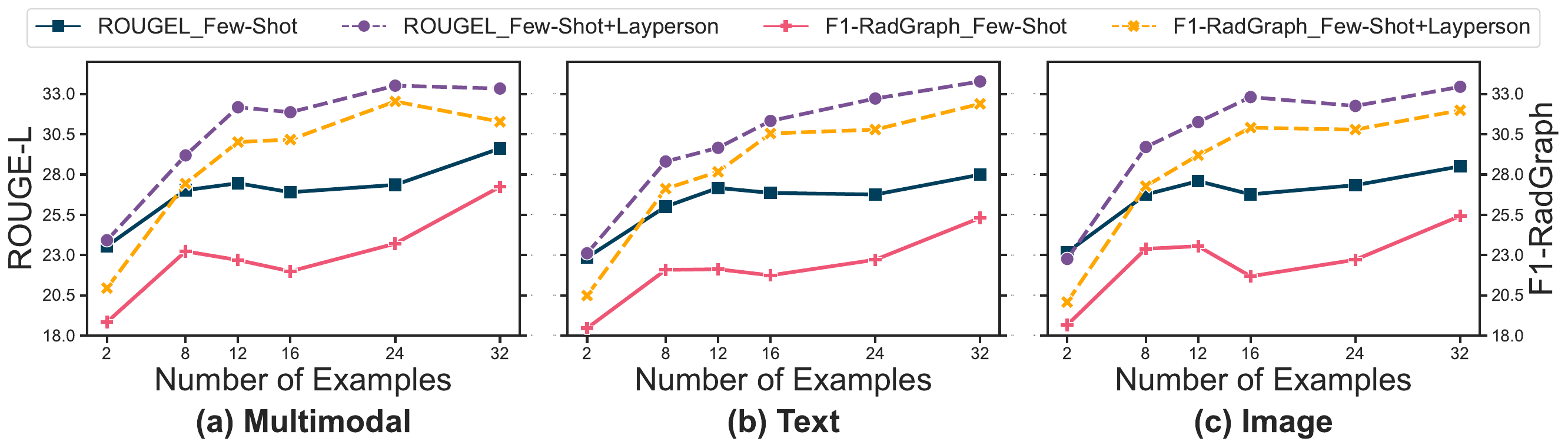}\par 
\vspace{-0.5em}
\caption{Validation results vs. the number of in-context examples across various prompt types and modality embeddings on Starling-LM-7B.
\label{fig:starling-plot}}
\end{figure*}

\begin{figure*}[htbp]
\centering
\includegraphics[width=\textwidth]{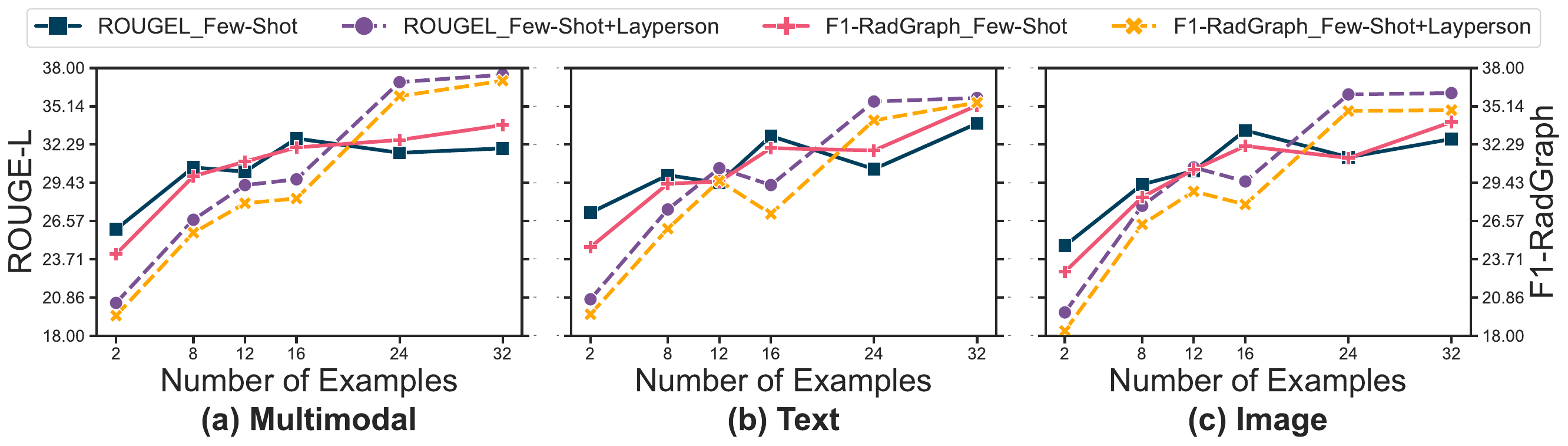}\par 
\vspace{-0.5em}
\caption{Validation results vs. the number of in-context examples across various prompt types and modality embeddings on Meta-Llama-3-8B-Instruct.
\label{fig:llama-plot}}
\end{figure*}

A natural question that arises is, ``Does integrating a larger number of examples in Few-Shot + Layperson prompting lead to better overall performance?''. To answer this question, we explore the relationship between performance and the number of examples integrated. 
To better quantify the contributions of different components in our model, we conducted ablation studies focusing on various prompt types and modality embeddings for the radiology reports summarization task. Using the MIMIC-CXR validation dataset, we evaluated the performance of three models, OpenChat-3.5-7B, Starling-LM-7B, and Meta-Llama-3-8B-Instruc across a range of configurations. Our analysis focuses on understanding the effectiveness of embedding matches for different modalities (including image, text, and multimodal), as well as determining the optimal number of examples needed for effective summarization. The results of these ablations on the MIMIC-CXR validation set are presented in Figure~\ref{fig:openchat-plot}, Figure~\ref{fig:starling-plot}, and Figure~\ref{fig:llama-plot}. Specifically, we note that Few-Shot + Layperson prompting with multimodal embedding matched examples slightly outperforms the image and text embedding matched ones. For all OpenChat-3.5-7B, Starling-LM-7B, and Meta-Llama-3-8B-Instruc employing the LaypersonPrompt demonstrates performance enhancements compared to the original prompt.

Furthermore, as we increase the number of examples, the performance continues to rise, which demonstrates that prompting the model with more in-context examples improves performance. 
However, we can also observe a slight performance decrease for in some cases after reaching 24 examples. These findings suggest that while multimodal embeddings provide a robust framework for summarization, there is a complex relationship between the number of examples and performance gains. Our studies highlight the importance of multimodal context and suggest a diminishing return for additional examples in text and image modalities beyond a certain point. This insight is critical for optimizing the efficiency and accuracy of our summarization model when processing radiology data.

\begin{table*}[htbp]
\centering
\resizebox{0.8\linewidth}{!}{
\begin{tabular}{@{}lrrrrrrr@{}}
\toprule
                              & \multicolumn{1}{l}{} & \textbf{2}    & \textbf{8}    & \textbf{12}   & \textbf{16}   & \textbf{24}   & \textbf{32}    \\ \midrule
\multirow{2}{*}{MIMIC-CXR} & \textcolor{black}{Few-Shot}           & 643  & 1285 & 1713 & 2141 & 2994 & 3850  \\
                              & \textcolor{black}{Few-Shot + Layperson}                & 889  & 1826 & 2452 & 3084 & 4333 & 5587  \\ \midrule
\multirow{2}{*}{MIMIC-III} & \textcolor{black}{Few-Shot}           & 1035 & 2500 & 3474 & 4451 & 6405 & 8359  \\
                              & \textcolor{black}{Few-Shot + Layperson}               & 1340 & 3277 & 4565 & 5856 & 8442 & 11025 \\ \bottomrule
\end{tabular}}
\caption{Average Token of Prompts.}
\label{tab:tokens}
\end{table*}

Table~\ref{tab:tokens} shows the prompt lengths corresponding to various numbers of examples used in our study. We aim to explore how the length of prompts affects model performance. Initially, we explored LLaMA-2-7B and GPT-Neo-2.7B. However, given that LLaMA-2-7B and LLaMA-2-13B has a maximum context length of 4,096 tokens and GPT-Neo-2.7B is restricted to 2,048 tokens, such constraints on context length impact the performance of LLaMA-2-7B, LLaMA-2-13B and GPT-Neo-2.7B in the radiology reports summarization task compared to models capable of processing longer contexts like OpenChat-3.5-7B and Starling-LM-7B (up to 8,192 tokens). Specifically, these constraints significantly affect LLaMA-2-7B, LLaMA-2-13B and GPT-Neo-2.7B's ability to conduct in-context learning for summarizing radiology reports. The restricted context length can hinder these models from fully taking advantage of the extensive information required for accurate summarization in this domain.

\textcolor{black}{Therefore, LLaMA-2-7B, LLaMA-2-13B and GPT-Neo-2.7B do not perform very well, which may be due to limitations in their reasoning capabilities and the constrained number of examples in few-shot learning scenarios, restricted by the maximum token count. This means that even if 16 examples are provided, the models may truncate the initial examples to stay within the token limit, potentially losing valuable context.}





\end{document}